\documentclass[11pt]{article}

\usepackage[final]{acl}

\usepackage{times}
\usepackage{latexsym}

\usepackage[T1]{fontenc}

\usepackage[utf8]{inputenc}

\usepackage{microtype}

\usepackage{inconsolata}

\usepackage{graphicx}
\usepackage{amsfonts}
\usepackage{booktabs}

\title{Attention Dispersion as a Diagnostic Signal for Hallucination in Large Language Models}

\author{Shardul P. More \\
  Rajarambapu Institute of Technology \\
  Ishwarpur, Maharashtra, India \\
  \texttt{shardulmore112@gmail.com} \\\And
  Tanuja S. Pawar \\
  Rajarambapu Institute of Technology \\
  Ishwarpur, Maharashtra, India \\
  \texttt{tanujapawar203@gmail.com} \\}

\begin{document}
\maketitle
\begin{abstract}
Large Language Models (LLMs) frequently exhibit hallucinations, presenting a major barrier to reliability in complex reasoning tasks. While traditional detection methods rely on output-based confidence metrics, these logits are often miscalibrated by modern alignment techniques. In this paper, we investigate the temporal volatility of internal attention mechanisms as an alternative diagnostic signal for hallucination that does not depend on output calibration. By introducing an unsupervised metric for attention dispersion, we show that epistemic uncertainty leaves a measurable trace within intermediate layers, where spikes in attention entropy are associated with reasoning breakdowns. We evaluate our approach on mathematical reasoning benchmarks (GSM8K and MATH-500) using the Qwen2.5 model family (1.5B and 3B parameters), finding statistically significant AUC improvements of up to +0.076 over output-based baselines across all tested conditions. These findings suggest that attention dispersion is a promising complement to traditional hallucination detection methods, requiring further investigation across broader model families and task domains.
\end{abstract}

\section{Introduction}
\label{sec:intro}

The rapid growth and integration of Large Language Models (LLMs) present a challenge in identifying hallucinations, in which the models are confident yet factually incorrect in their outputs. Initial approaches to uncertainty estimation heavily relied on the model's output probabilities, leveraging maximum softmax probabilities as a baseline for detecting errors \citep{hendrycks_baseline_2018}. Further studies demonstrated that large language models can often evaluate the validity of their own claims through self-evaluation prompts \citep{kadavath_language_2022}. However, recent alignment strategies, such as Reinforcement Learning from Human Feedback (RLHF), often cause models to exhibit systematic, verbalized overconfidence regardless of actual response quality \citep{leng_taming_2025}. Consequently, the output logits are structurally compromised as diagnostic indicators.

Recent methodologies have explored internal representations to bypass the corrupted logits. Classifiers such as SAPLMA \citep{azaria_internal_2023} are used to extract signals of truthfulness directly from hidden-layer activations. More importantly, these approaches often do not consider the dynamic, multi-step nature of sequential reasoning. Even though mid-depth layers act as critical information bottlenecks \citep{skean_layer_2025}, the temporal fluctuations or volatility of the attention mechanism across these layers has not been thoroughly investigated as a diagnostic tool.

We hypothesize that hallucination is not just a static representation error, but a dynamic failure in reasoning stability. Grounded reasoning maintains a consistent attention trajectory through the depth of the transformer, whereas epistemic failure appears as erratic, highly dispersed attention shifts. Our contributions are threefold: (1) we propose the theoretical framing that the temporal dispersion of attention entropy can serve as a zero-shot, internal indicator of epistemic failure; (2) we introduce a metric, $step\_attn\_std$, to track reasoning volatility across intermediate layers without relying on output logits; and (3) we provide empirical validation on GSM8K and MATH-500, demonstrating that dynamic attention dispersion provides a predictive signal for hallucination detection without requiring additional probe training.

\section{Related Work}
\label{sec:related_work}
For the purpose of enhancing the reliability of the reasoning of large language models and measuring their uncertainty, researchers have explored sampling-based methods. Self-Consistency was introduced as a method that selects the most frequently occurring answer among diverse reasoning paths, showcasing significant improvements in logical as well as general commonsense tasks \citep{wang_self-consistency_2023}. Similarly, Semantic Entropy was proposed to cluster multiple responses based on semantic equivalence, detecting hallucinations by measuring uncertainty across these clusters \citep{farquhar_detecting_2024}. While these strategies provide robust uncertainty signals, they require multiple generation passes per input. For instance, Semantic Entropy requires 5 to 10 generations, resulting in higher computational costs. To mitigate this overhead, recent work has proposed Semantic Entropy Probes (SEPs), which approximate semantic uncertainty directly from the hidden states of a single generation \citep{kossen_semantic_2024}.

Another line of research is verbalized confidence, where the language models are required to report the certainty level of a generated answer explicitly. Studies on the CalibratedMath benchmark showed that models can be trained to express uncertainty by reporting numerical probabilities or linguistic features describing something \citep{lin_teaching_2022}. Furthermore, confidence elicitation has been expanded to RLHF-aligned models, discovering that verbalized confidence can sometimes offer better calibration than the model's conditional probabilities \citep{tian_just_2023}. However, while models can meaningfully estimate confidence in natural language, this signal remains dependent on the model's explicit verbal outputs, which can still be likely influenced by alignment-induced overconfidence \citep{leng_taming_2025}.

To bypass output-dependent metrics, mechanistic interpretability analyzes internal model computations. Previous work has mapped Transformer circuits to understand component interactions \cite{elhage_mathematical_2021} and identified middle-layer feed-forward networks as essential mediators of factual prediction \cite{meng_locating_2022}. These studies confirm that robust knowledge representations reside strictly within intermediate layers. Building on this foundation, we shift from static feature extraction to dynamically tracking the temporal instability of these attention mechanisms.
\section{Methodology}
\label{sec:methodology}

Our objective is to quantify the knowledge-related stability of a Large Language Model (LLM) during sequential reasoning without relying on its output logits. To achieve this, we introduce a framework that measures the temporal dispersion of attention entropy across the network's layers at each generation step.

\subsection{Layer-Wise Attention Entropy}
Consider a transformer model with $L$ layers. At generation step $t$, the model attends to the preceding context of length $k$. For a given layer $l \in \{1, \dots, L\}$ and attention head $h \in \{1, \dots, H\}$, let $\alpha_{l,h}^{(t)} \in \mathbb{R}^k$ denote the attention probability distribution over the context tokens.

To quantify the positional or spatial focus of the model at layer $l$, we first calculate the Shannon entropy of the attention distribution. We define the layer-wise attention entropy $E_l^{(t)}$ as the average entropy across all $H$ heads in layer $l$:

$$E_l^{(t)} = \frac{1}{H} \sum_{h=1}^{H} \left( - \sum_{i=1}^{k} \alpha_{l,h,i}^{(t)} \log \alpha_{l,h,i}^{(t)} \right)$$

A low entropy value indicates a sharp, highly localized attention focus, while a high entropy value indicates broad, dispersed attention across the context.

\subsection{Temporal Attention Dispersion}
Because intermediate layers systematically compress semantic features \citep{skean_layer_2025}, grounded reasoning should exhibit a well-connected, stable trajectory of attention entropy through this depth. In contrast, we propose that epistemic failure disrupts this trajectory, causing volatile shifts in attention entropy across layers.

To measure this volatility, we calculate the standard deviation of the layer-wise entropies across the entire depth of the model. First, we define the mean attention entropy across all layers at step $t$:

$$\bar{E}^{(t)} = \frac{1}{L} \sum_{l=1}^{L} E_l^{(t)}$$

Next, we define our primary metric, temporal attention dispersion ($step\_attn\_std$), as the standard deviation of these layer-wise entropies from the mean:

$$step\_attn\_std^{(t)} = \sqrt{ \frac{1}{L} \sum_{l=1}^{L} \left( E_l^{(t)} - \bar{E}^{(t)} \right)^2 }$$

By extracting $step\_attn\_std^{(t)}$ at each generation step, we obtain a dynamic, zero-shot signal of the model's reasoning stability. This metric is computed strictly from the internal self-attention mechanisms, rendering it completely independent of miscalibrated output logits.

\section{Experiments}

\label{sec:experiments} 

To test our hypothesis that temporal attention dispersion is a reliable signal of epistemic failure, we evaluate our metrics against standard output-based measures of confidence across multiple reasoning datasets and model scales.

\subsection{Datasets and Models}
We select two complex mathematical reasoning benchmarks that require multi-step sequential logic, where output-level hallucination is highly prevalent: GSM8K \citep{cobbe_training_2021}, a dataset of high-quality grade-school math word problems, and MATH-500 \citep{lightman_lets_2023}, a curated subset of the MATH dataset featuring highly complex, competition-level mathematics.

To evaluate the scalability and robustness of our approach, we extract internal states and outputs from two different models of the Qwen family\citep{yang_qwen2_2024}: a 1.5B-parameter model (evaluated on both GSM8K and MATH-500) and a 3B-parameter model (evaluated on MATH-500). Ground-truth labels are extracted using a strict, stack-based LaTeX parsing algorithm to perfectly align model generations with boxed mathematical solutions.

\subsection{Feature Extraction}
For each generated trajectory, we aggregate the step-level metrics into sequence-level features to train a lightweight hallucination classifier. For \textbf{output baseline features}, we compute the mean and maximum of the predictive entropy at the output layer ($mean\_out\_entropy$, $max\_out\_entropy$), alongside the sequence-level output probability ($out\_tau$). For our proposed \textbf{attention features}, we aggregate the spatial attention entropy ($mean\_attn\_entropy$, $max\_attn\_entropy$) and sequence-level attention confidence ($attn\_tau$). Crucially, we also include the temporal dispersion of attention across layers, aggregated over the sequence ($mean\_attn\_std$).

\subsection{Evaluation Protocol}

We frame hallucination detection as a binary classification task where the objective is to predict reasoning failure (i.e., whether the final generated answer is incorrect). We train a standard Logistic Regression classifier using Scikit-learn \cite{pedregosa_scikit-learn_2018} with balanced class weights to predict failure based on the extracted features. 

To ensure robust estimation of model performance and prevent overfitting, we employ a Repeated Stratified K-Fold cross-validation strategy (5 splits, 10 repeats). All features are standardized prior to training. We evaluate the models using the Area Under the Receiver Operating Characteristic Curve (ROC-AUC). To determine statistical significance between the proposed attention-based features and the output-based baselines, we compute 95\% confidence intervals for the performance delta and apply a one-sided Wilcoxon signed-rank test.

\section{Results and Analysis}
\label{sec:results}

\begin{figure*}[t]
    \centering
    \begin{minipage}{0.48\textwidth}
        \centering
        \includegraphics[width=\linewidth]{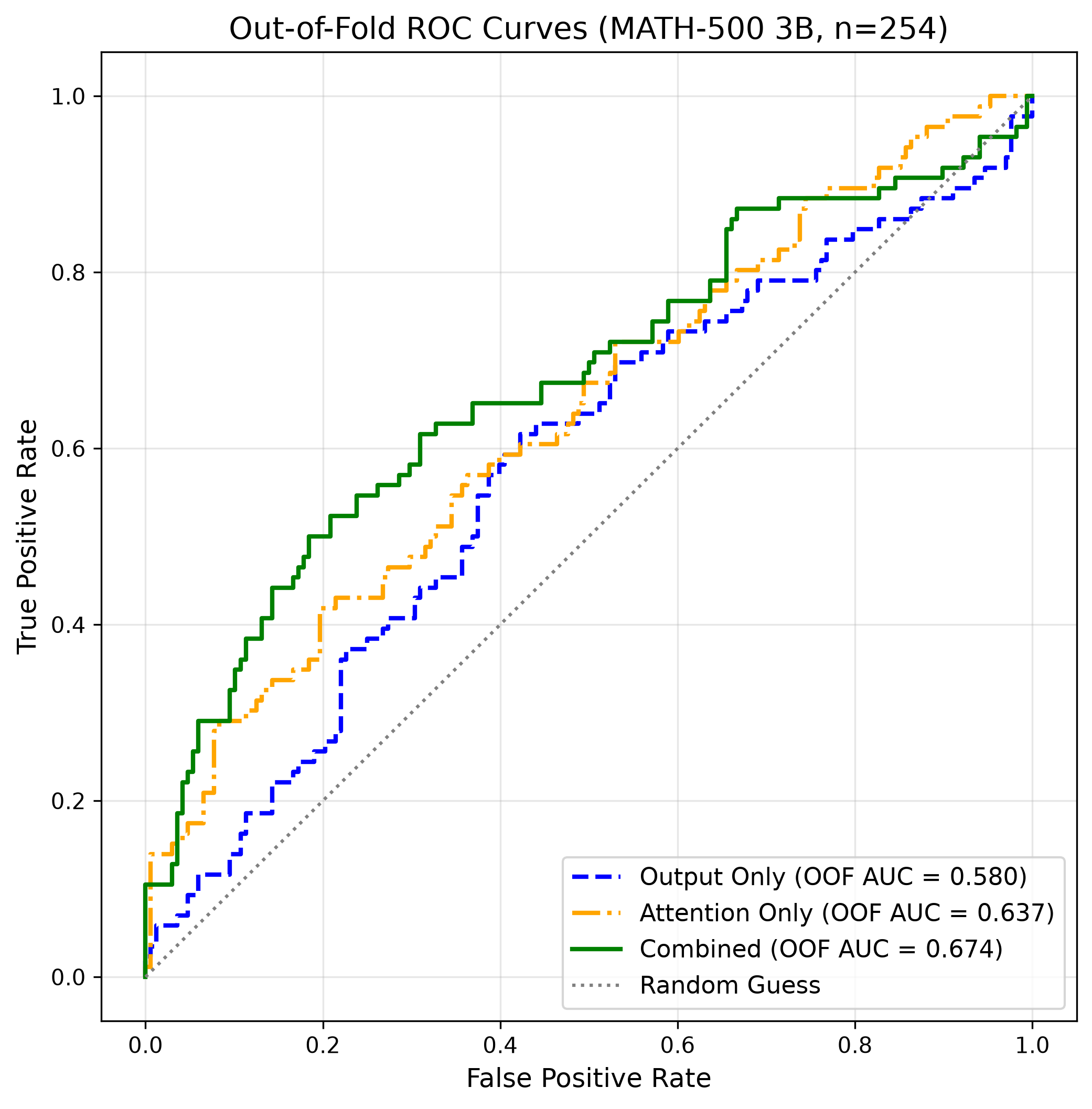}
        \caption{Out-of-Fold ROC Curves on MATH-500 (3B). The internal attention features outperform standard output metrics, while the combined model achieves the highest AUC.}
        \label{fig:roc_curves}
    \end{minipage}\hfill
    \begin{minipage}{0.48\textwidth}
        \centering
        \includegraphics[width=\linewidth]{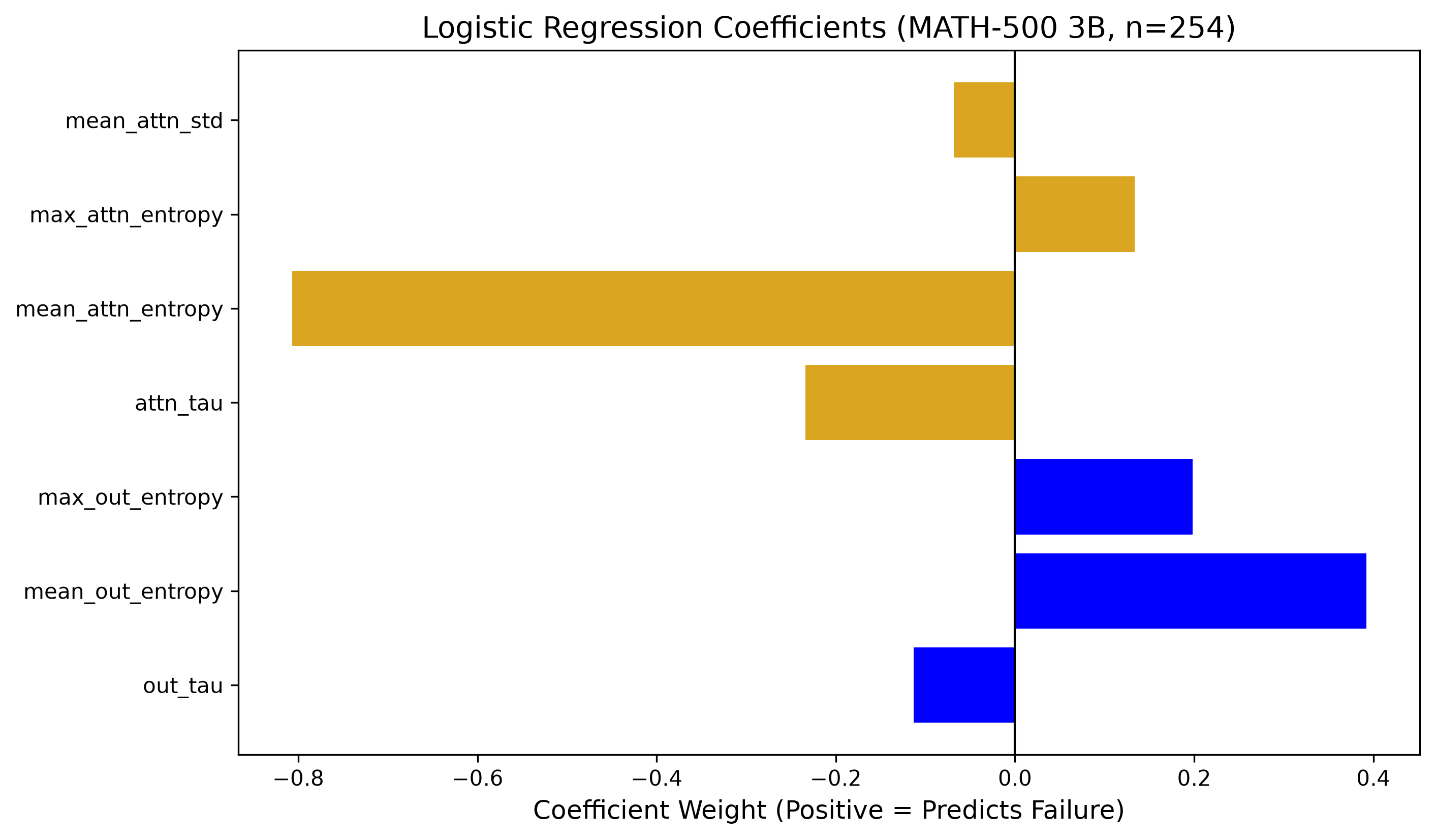}
        \caption{Logistic Regression Coefficients on MATH-500 (3B). Positive weights strongly predict reasoning failure.}
        \label{fig:coefficients}
    \end{minipage}
\end{figure*}

Our evaluations suggest that internal attention dispersion provides a predictive, orthogonal signal for hallucination detection that consistently outperforms traditional output-based metrics on complex reasoning tasks.

\subsection{Performance on Reasoning Benchmarks}
Table \ref{tab:main_results} summarizes the classification performance across different benchmarks and model scales. The proposed attention-driven feature set consistently outperforms the baseline across all tested configurations. 

Across the evaluated settings, attention dispersion appears increasingly informative in larger models and more challenging reasoning tasks. However, these observations are suggestive rather than conclusive, as model scale and reasoning complexity are not independently controlled in our experiments. While the proposed metric yields a modest gain on the GSM8K benchmark (+0.0093 AUC), the performance gain is larger on the more challenging MATH-500 benchmark, reaching +0.0486 AUC for the 1.5B model and +0.0759 AUC for the 3B model.

\begin{table*}[t]
    \centering
    \small 
    \setlength{\tabcolsep}{4pt} 
    \renewcommand{\arraystretch}{0.9} 
    \begin{tabular}{llcccc}
        \toprule
        \textbf{Benchmark} & \textbf{Model} & \textbf{Base AUC} & \textbf{Prop. AUC} & \textbf{Gain (95\% CI)} & \textbf{$p$-value} \\
        \midrule
        GSM8K & Qwen2.5-1.5B & 0.6897 & 0.6990 & +0.0093 ([0.003, 0.015]) & $1.66 \times 10^{-3}$ \\
        MATH-500 & Qwen2.5-1.5B & 0.6259 & 0.6745 & +0.0486 ([0.031, 0.067]) & $5.49 \times 10^{-6}$ \\
        MATH-500 & Qwen2.5-3B & 0.5817 & 0.6575 & +0.0759 ([0.057, 0.095]) & $5.65 \times 10^{-8}$ \\
        \bottomrule
    \end{tabular}
    \vspace{-2mm} 
    \caption{Evaluation results showing AUC performance across models and datasets. The 95\% confidence intervals (CI) and $p$-values (Wilcoxon signed-rank test) confirm the statistical significance of the gains.}
    \label{tab:main_results}
\end{table*}

As further illustrated in Figure \ref{fig:roc_curves}, the fusion of both signals (Combined) yields the highest predictive power, indicating that temporal attention dispersion captures failure modes not well reflected in output logits alone.

\subsection{Mechanistic Feature Importance}

To understand the drivers of this performance, we analyze the feature weights of the trained logistic regression classifier. The analysis reveals that $mean\_attn\_entropy$ and $max\_attn\_entropy$ carry significant weight, but critically, they act in opposing directions. A high maximum attention entropy predicts failure, aligning with our hypothesis that sudden, dispersed attention spikes indicate a breakdown in reasoning focus.

The classifier's reliance on these internal attention features, even when output metrics like $out\_tau$ are available, suggests that attention volatility carries information about hallucination risk not captured by output confidence alone --- a pattern consistent with our hypothesis, though further work across model families and task types is needed to establish generality.

\section{Conclusion}
\label{sec:conclusion}
We investigated the temporal volatility of internal attention mechanisms as a diagnostic signal for hallucination in LLMs. Our results suggest that epistemic uncertainty leaves a measurable trace within intermediate layers, offering a signal less exposed to the miscalibration affecting output logits. Using an unsupervised metric for attention dispersion, we found that spikes in attention entropy are associated with reasoning breakdowns. Evaluations on GSM8K and MATH-500 with the Qwen2.5 family show this internal signal consistently outperforms output-based baselines across all tested conditions, with gains that appear larger in more challenging configurations — though model scale and task complexity are not independently controlled in our design. These findings point toward internal computational signals as a promising complement to output-based hallucination detection. Future work will extend this framework to diverse architectures, non-mathematical domains, and direct comparison against sampling-based uncertainty methods, while exploring the causal drivers of these attention breakdowns.

\section*{Limitations}
\label{sec:limitations}

While our findings suggest the potential of temporal attention dispersion as a diagnostic measure for hallucination, this study has several important limitations that require consideration:

\textbf{Domain Specificity:} Our empirical evaluation is strictly confined to mathematical reasoning tasks (GSM8K and MATH-500). While multi-step mathematics provides an excellent testbed for sequential logic, the observed attention dynamics need to be analyzed in other modalities of hallucination, such as factual fabrications in open-domain question answering, creative generation, or translation tasks.

\textbf{Architectural Scope:} The experiments were conducted solely on the Qwen2.5 model family (1.5B and 3B parameters). We have not yet verified whether these temporal attention dispersion patterns remain consistent across fundamentally different Transformer variants, such as Mixture-of-Experts (MoE) architectures, or models utilizing alternative attention mechanisms.

\textbf{White-Box Requirement:} Our proposed metric, $step\_attn\_std$, fundamentally relies on extracting intermediate attention distributions across all layers of the network. Consequently, this diagnostic framework is limited to open-weight models and cannot be applied to proprietary, black-box APIs (e.g., GPT-4, Claude) where only final textual outputs or restricted output logits are accessible. 

\textbf{Confounding Variables in Scaling:} As noted in our analysis, the predictive advantage of our metric appeared greater on the larger model (3B) and the more complex benchmark (MATH-500). However, our experimental design did not independently control for model scale versus task complexity, limiting our ability to determine what is actually driving this trend.

\textbf{Missing Empirical Baselines:} While we position our work against sampling-based uncertainty methods such as Semantic Entropy and Semantic Entropy Probes in our related work, we do not include a direct empirical comparison against these methods in this study. Our evaluation is limited to output-logit-based baselines; establishing relative performance against these alternative approaches is left to future work.

\textbf{Sample Size:} Our evaluation, particularly for the 3B model on MATH-500 ($n=254$), involves a limited sample size relative to the scale often used in large-scale uncertainty quantification studies. While our statistical tests indicate significant effects within this sample, further validation on larger evaluation sets would strengthen confidence in the generalizability of the reported effect sizes.

\textbf{Step Alignment Coverage:} Our step-level parsing algorithm did not achieve complete alignment coverage on MATH-500 due to its more complex LaTeX notation compared to GSM8K, potentially introducing selection effects if parsing failures are not independent of reasoning correctness.

\section*{Acknowledgments}
This research was conducted independently without external institutional funding. The authors gratefully acknowledge Kaggle for providing the computation required for the execution of the model evaluations. We disclose the use of generative AI tools during manuscript and pipeline preparation: Anthropic's Claude Sonnet 5 was used to assist with code generation and pipeline development, and Google's Gemini 3.1 Pro and Grammarly were used for proofreading and language polishing. All AI-assisted code, analysis, and text were reviewed, verified, and revised by the authors, who take full responsibility for the content of this manuscript.


\bibliography{custom}

\end{document}